\documentclass[10pt,twocolumn,letterpaper]{article}

\usepackage{iccv}
\usepackage{times}
\usepackage{epsfig}
\usepackage{graphicx}
\usepackage{amsmath}
\usepackage{multirow}
\graphicspath{{figure-main/}} 
\usepackage{url} 
\usepackage{colortbl}

\def\eg{\emph{e.g.}} 

\def\ie{\emph{i.e.}}

\def\etc{\emph{etc.}}

\usepackage[pagebackref=true,breaklinks=true,letterpaper=true,colorlinks,bookmarks=false]{hyperref}

\iccvfinalcopy 

\def\iccvPaperID{8484}

\ificcvfinal\fi

\begin{document}

\title{Webly Supervised Fine-Grained Recognition:\\Benchmark Datasets and An Approach}

\author{Zeren Sun$^{1}$,
		Yazhou Yao$^{1}$\thanks{Corresponding author.},
		Xiu-Shen Wei$^{1,2 *}$, 
		Yongshun Zhang$^2$,\\ 
		Fumin Shen$^3$,
		Jianxin Wu$^2$,
		Jian Zhang$^4$, 
		Heng Tao Shen$^3$\\
$^1$School of Computer Science and Engineering, Nanjing University of Science and Technology\\
$^2$State Key Laboratory for Novel Software Technology, Nanjing University\\
$^3$University of Electronic Science and Technology of China \ \ \ 
$^4$University of Technology Sydney\\
}

\maketitle
\ificcvfinal\thispagestyle{empty}\fi

\begin{abstract}
Learning from the web can ease the extreme dependence of deep learning on large-scale manually labeled datasets. Especially for fine-grained recognition, which targets at distinguishing subordinate categories, it will significantly reduce the labeling costs by leveraging free web data. Despite its significant practical and research value, the webly supervised fine-grained recognition problem is not extensively studied in the computer vision community, largely due to the lack of high-quality datasets. To fill this gap, in this paper we construct two new benchmark webly supervised fine-grained datasets, termed WebFG-496 and WebiNat-5089, respectively. In concretely, WebFG-496 consists of three sub-datasets containing a total of 53,339 web training images with 200 species of birds (Web-bird), 100 types of aircrafts (Web-aircraft), and 196 models of cars (Web-car). For WebiNat-5089, it contains 5089 sub-categories and more than 1.1 million web training images, which is the largest webly supervised fine-grained dataset ever. As a minor contribution, we also propose a novel webly supervised method (termed ``{Peer-learning}'') for benchmarking these datasets.~Comprehensive experimental results and analyses on two new benchmark datasets demonstrate that the proposed method achieves superior performance over the competing baseline models and states-of-the-art. Our benchmark datasets and the source codes of Peer-learning have been made available at {\url{https://github.com/NUST-Machine-Intelligence-Laboratory/weblyFG-dataset}}.

\end{abstract}

\section{Introduction}

Recent success of deep learning has shown that a deep network in conjunction with abundant well-labeled training data is the most promising approach for fine-grained recognition~\cite{wei2018mask,complementarypart2019,guosencvpr2019,guoseneccv2020}. 
However, even with the availability of scalable crowd-sourcing platforms like Amazon Mechanical Turk, constructing a large-scale fine-grained dataset like iNat2017~\cite{van2018inaturalist} is still an extremely difficult work since distinguishing subtle differences among fine-grained categories (\eg, different animals~\cite{nabirds,dogseccv2012}, or plants~\cite{nilsback2006visual,Hou2017VegFru}) usually requires domain-specific expert knowledge.

To reduce the cost of manual fine-grained annotation, many methods have been proposed, which are primarily focused on semi-supervised learning~\cite{cui2016fine}. These works inevitably involve various forms of human intervention and remain labor-consuming \cite{yao2021non}. To further reduce manual annotations as well as to learn more practical fine-grained models, training directly from web images is becoming increasingly popular~\cite{zhang2021mm,yao2021jo,liutmm2021,yao2020bridging}. Nevertheless, the lacking of a benchmark dataset makes it difficult to fairly compare the performances of these algorithms. This is the motivation of this work and we aim to provide a benchmark dataset for evaluating webly supervised fine-grained recognition algorithms. 

\begin{figure*}[t]
	\centering
	\includegraphics[width=0.98\textwidth]{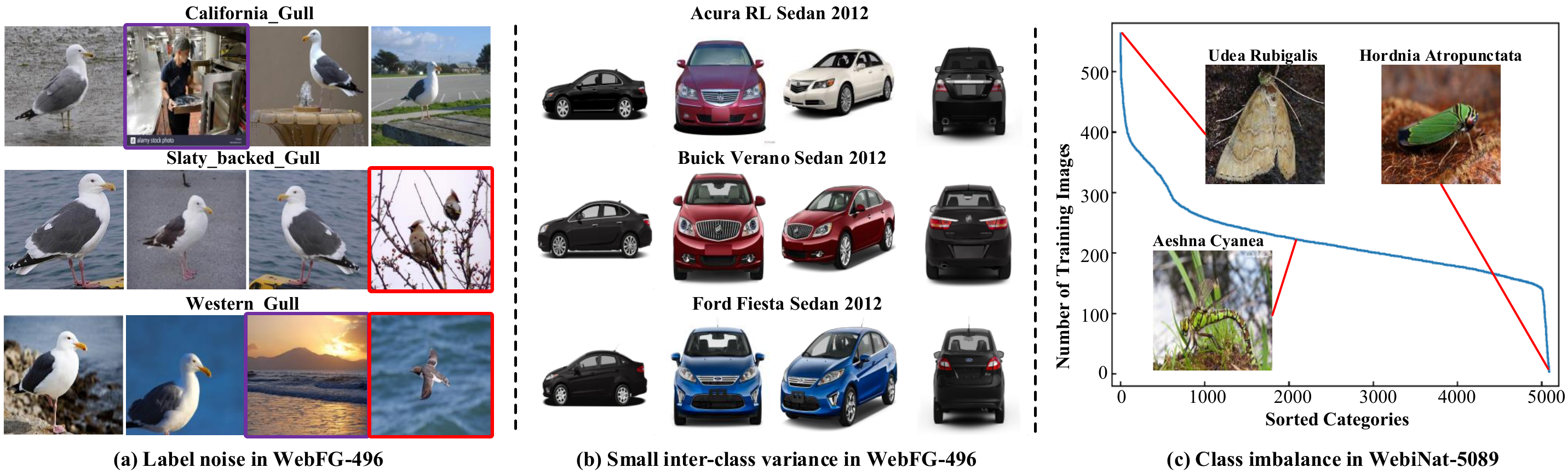}
	\caption{\small{Examples of WebFG-496 and WebiNat-5089. (\textbf{a}) Cross-domain (purple bounding box) and cross-category (red bounding box) noises in WebFG-496. (\textbf{b}) Due to the variety of colors, poses and other factors, there are small inter-class and large intra-class variances among subcategories in the WebFG-496 dataset. (\textbf{c}) Distribution of training images per category in WebiNat-5089. The number of training images in the largest subcategory ``Udea Rubigalis'' is 563 while in the smallest subcategory ``Hordnia Atropunctata'' is only 4.}}
	\label{fig1}
	\vspace{-0.2cm}
\end{figure*}

Webly supervised fine-grained recognition consists of three challenges: (1) \textbf{Label noise} - Different from manually labeled datasets, fine-grained web images are often associated with label noises due to the influence of error-prone automatic or non-domain-expert annotations. We consider two types of label noises, which we call ``cross-domain'' and ``cross-category'' noises. To be specific, cross-domain noise is the portion of images that are not of any categories in the same fine-grained domain, \eg, for birds, it is the fraction of images that do not contain a bird (cf. images with purple bounding boxes in Fig.~\ref{fig1}~(a)). In contrast, cross-category noise is the portion of images that have the wrong labels within a fine-grained class, \eg,~an image of a bird with the wrong category label (cf. images with red bounding boxes in Fig.~\ref{fig1}~(a)).
According to existing works~\cite{motivation2017}, deep neural networks have memorization effects, which will memorize incorrectly labeled training data and cause a poor generalization performance~\cite{cheniccv2015}. 
(2) \textbf{Small inter-class variance} - As shown in Fig.~\ref{fig1}~(b), three fine-grained sub-categories have small inter-class variance while each sub-category has large intra-class variance. Aiming at recognizing hundreds of sub-categories belonging to the same super-category is an extremely challenging task.
(3) \textbf{Class imbalance} - The natural world is heavily imbalanced, as some species are much more likely to be observed~\cite{van2018inaturalist}. As shown in Fig.~\ref{fig1}~(c), we collected 563 web training images for ``Udea Rubigalis''. While for ``Hordnia Atropunctata'', we only gathered 4 web images for training. Extreme class imbalance is a property of the real world fine-grained categories and recognition models should can deal with it.

To address aforementioned issues, we leverage the categories in three famous fine-grained datasets, \ie, FGVC-Aircraft~\cite{maji13fine-grained}, CUB200-2011~\cite{cub200-2011} and Stanford Cars~\cite{KrauseStarkDengFei-Fei_3DRR2013}, to construct a new webly supervised fine-grained benchmark dataset WebFG-496. To scale up the categories as well as to build a much larger and challenging fine-grained dataset, we reuse the categories in iNat2017~\cite{van2018inaturalist} to build a web version iNaturalist dataset, termed as WebiNat-5089. Compared to manually labeled iNat2017, WebiNat-5089 consists of 1,184,520 web training images, basically twice the number of training images (\ie, 579,184) in the original iNat2017. 

Furthermore, we also propose a simple yet effective learning paradigm to train robust deep fine-grained models directly from noisy web images. Our work is motivated by the following observations: {\textbf{1)}} The performance of deep neural networks can be boosted by learning from wrongly classified instances (\ie, ``hard examples'')~\cite{decoulping}. {\textbf{2)}} Deep neural networks always fit to ``easy examples'' first, and gradually adapt to ``hard examples''~\cite{motivation2017}. {\textbf{3)}} Distinct networks have different learning abilities, and can collaboratively boost their performance by mutually communicating ``useful information''~\cite{coteaching}. Specifically, we train two deep neural networks simultaneously and let them mutually correct their classification errors. For each mini-batch of web images, each network individually feeds forward all data to separately predict the labels, based on which the input data is split into two groups $G_d$ (instances with different predictions) and $ G_s $ (instances with identical predictions). Then, the networks update their parameters with the instances in $G_d$. Meanwhile, each network sorts and fetches small-loss instances in $ G_s $ as ``useful knowledge'', and communicates this ``useful knowledge'' with its peer network for correcting the errors induced by instances in $ G_d $ during updating. Extensive experiments on the new benchmark datasets demonstrate the effectiveness of our proposed approach. The main contributions of this work can be summarized as follows:

	
\noindent \textbf{(1)} 
We construct two webly supervised fine-grained datasets, \ie, WebFG-496 and WebiNat-5089, for fairly benchmarking webly supervised fine-grained approaches. Specifically, WebFG-496 consists of three sub-datasets: Web-aircraft, Web-bird, and Web-car. It can help researchers promptly verify the effectiveness of their proposed methods. The large-scale WebiNat-5089 contains 5089 fine-grained subcategories and more than 1.1 million web training images. To the best of our knowledge, it is the largest webly supervised fine-grained benchmark dataset.
	
\noindent \textbf{(2)} 
We propose a novel deep learning paradigm, \ie, Peer-learning for dealing with webly supervised fine-grained recognition. Our model jointly leverages both ``hard'' and ``easy'' examples for training, which can keep the peer networks diverged and maintain their distinct learning abilities in removing noisy images.~This strategy can alleviate both the accumulated error problem in MentorNet~\cite{mnet} and the consensus issue in Co-teaching~\cite{coteaching}, which thus can boost the performance of webly supervised learning. 
	
\noindent \textbf{(3)}
We conduct extensive experiments of various baseline methods to benchmark the proposed WebFG-496 and WebiNat-5089 datasets, as well as our Peer-learning. In experiments, results on WebFG-496 and WebiNat-5089 validate the superiority of our method over states-of-the-arts, and ablation studies justify both the merits of the proposed benchmark datasets and the effectiveness of our method.
	

\section{Related Work}

\subsection{Fine-Grained Recognition Datasets}

In the past decade, the computer vision community has developed many fine-grained datasets covering diverse meta-categories, \eg, aircrafts~\cite{maji13fine-grained,airplane14}, birds~\cite{cub200-2011,birdsnap}, cars~\cite{KrauseStarkDengFei-Fei_3DRR2013,careccv14,carcvpr15,caraaai17}, dogs~\cite{dogseccv2012,khosla2011novel,parkhi12a}, flowers~\cite{nilsback2006visual}, food~\cite{Hou2017VegFru}, leaves~\cite{leaves}, trees~\cite{trees}, insects~\cite{van2018inaturalist}, \etc. The statistics of popular used fine-grained datasets are provided in Table~\ref{tab1}. Compared to coarse-grained annotations, fine-grained image labeling is an extremely difficult task and only a small number of domain experts are capable of correctly labeling them. This motivates the requirement to automatically learning fine-grained recognition models from \emph{free} web images for discriminating large numbers of potentially visual similar categories. However, existing webly supervised datasets, like NUS-WIDE~\cite{chua2009nus}, WebVision~\cite{webvision} and OpenImages~\cite{Openimages}, are all coarse-grained datasets. Benchmarked webly supervised datasets tailored for fine-grained are required to fairly evaluate the performance of proposed approaches. Our two benchmark webly supervised fine-grained datasets, \ie, WebFG-496 and WebiNat-5089, are built in this context, which can promote further studies in this learning scenario.

\subsection{Webly Supervised Learning}

Training fine-grained recognition models with web images usually results in poor performance due to the presence of label noises and data bias \cite{sun2020crssc,zhang2020data,yao2020exploiting}. Statistical learning has contributed significantly to solve this problem, especially in theoretical aspects. In this work, our focus is on deep learning-based approaches. Roughly speaking, these works can be separated into four groups. The first group involves developing novel loss functions~\cite{yao2016domain,yao2019tmm,yao2019tip,reed2014training,flatow2017robustness,pencil2019} for dealing with label noises. The second group tries to estimate the noise transition matrix~\cite{goldberger2017,patrini}. The third one applies attention mechanisms to alleviate noises and data bias~\cite{zhuang2017attend}. The last group attempts to clean the web data as a preprocessing step~\cite{decoulping,coteaching,zhang2020aaai,yao2017exploiting,mnet}. However, none of these works are specifically designed for fine-grained visual recognition. Our proposed Peer-learning method belongs to the last group and is proposed for webly supervised fine-grained recognition. Different from existing methods (\eg, Decoupling~\cite{decoulping} which solely uses ``hard examples'', MentorNet~\cite{mnet} and Co-teaching~\cite{coteaching} which only explore ``easy examples'') our approach leverages both ``easy'' and ``hard'' examples for cross-updating two networks, can alleviate the accumulated errors and consensus issues in training webly supervised fine-grained recognition models. 

\begin{table}[t]\scriptsize{
		\centering
		\renewcommand{\arraystretch}{1.1}
		\caption{The statistics of popular fine-grained datasets. ``Supervision'' means the training data is manually labeled (``Manual'') or collected from the web (``Web'').}
		\vspace{0.1cm}
		\begin{tabular}{c|r|r|c|c}
			\hline
			\textbf{Domains}	&  \textbf{Dataset Name\:\:\:\:}                   &  \textbf{\# Train\:\:}          &  \textbf{\# Classes}  &  \textbf{Supervision}  \\
			\hline
			\multirow{2}{*}{Birds}  
			&  CUB200-2011~\cite{cub200-2011}             & 5,994                       & 200                      & Manual         \\
			&  NABirds~\cite{nabirds}                     & 23,929                      & 555                      & Manual         \\
			
			\hline
			\multirow{2}{*}{Cars}      
			&  Stanford Cars~\cite{KrauseStarkDengFei-Fei_3DRR2013} & 8,144             & 196                      & Manual         \\
			&  Census Cars~\cite{caraaai17}              & 512,765                      & 2,675                    & Manual         \\
			\hline
			\multirow{3}{*}{Dogs}
			&  Stanford Dogs~\cite{khosla2011novel}      & 12,000                       & 120                      & Manual         \\
			&  Oxford Pets~\cite{parkhi12a}              & 3,680                        & 37                       & Manual         \\
			&  DogSnap~\cite{dogseccv2012}               & 4,776                        & 133                      & Manual         \\  
			\hline
			\multirow{1}{*}{Aircraft}		
			&  FGVC-Aircraft~\cite{maji13fine-grained}   & 3,334                        & 100                      &  Manual        \\
			\multirow{1}{*}{Flowers}   
			& Flowers 102~\cite{nilsback2006visual}      & 1,020                        & 102                      & Manual         \\
			\multirow{1}{*}{Food}   
			& VegFru~\cite{Hou2017VegFru}                & 29,200                       & 292                      & Manual         \\
			\multirow{1}{*}{Leaves}   
			& LeafSnap~\cite{leaves}                     & 23,147                       & 185                      & Manual         \\
			\multirow{1}{*}{Trees}   
			& Urban Trees~\cite{trees}                   & 14,572                       & 18                       & Manual         \\
			\multirow{1}{*}{Natural}   
			& iNat2017~\cite{van2018inaturalist}         & 579,184                      & 5,089                    & Manual         \\
			\hline
			\multirow{2}{*}{\textbf{Ours}}   
			& \textbf{WebFG-496}                         & 53,339                       & 496                      & Web            \\
			& \textbf{WebiNat-5089}                      & 1,184,520                    & 5,089                    & Web            \\
			\hline
		\end{tabular}\\
		\label{tab1}}
	\vspace{-1em}
\end{table}  

\section{Datasets Construction}

In this section, we explain the construction details of our webly supervised fine-grained datasets WebFG-496 and WebiNat-5089. It should be noted that WebFG-496 consists of three sub-datasets: Web-aircraft, Web-bird, and Web-car. 

\begin{table*}[t]\small
	\centering
	\renewcommand{\arraystretch}{1.1}
	\caption{\small{Detailed construction process of training data in WebFG-496 and WebiNat-5089. ``Testing Source'' indicates where testing images come from. ``Imbalance'' is the number of images in the largest class divided by the number of images in the smallest.}}
	\vspace{0.1cm}
	\resizebox{17.5cm}{!}{
		\begin{tabular}{c|c|c|c|c|c|c|c}
			\hline
			
			\textbf{Dataset} & \textbf{Sub-dataset} & \textbf{Classes}  & \textbf{Testing Source} & \textbf{Retrieved Images} & \textbf{After Broken Filtering} & \textbf{After Duplicated Filtering} & \textbf{Imbalance} \\
			\hline
			\multirow{3}{*}{WebFG-496} 
			&  Web-aircraft            &  100  & Airliners                             &  14,817   &  14,772  & 13,503    & 1.2    \\
			
			& Web-bird\:\:\:\:\:       &  200  & Flickr                                &  29,211   &  29,098  & 18,388    & 1.1    \\
			
			& Web-car\:\:\:\:\:        &  196  & GIS+ Flickr+BIS &  27,959   &  27,895  & 21,448    & 2.3    \\
			\hline 
			WebiNat-5089               & -     &  5,089                   & iNaturalist & 1,437,483  & 1,434,083 & 1,184,520 & 140.8  \\
			\hline  	    	    	    	    	    	     	       						
	\end{tabular}}\\
	\label{tab2}
	\vspace{-1em}
\end{table*}

\textbf{Fine-Grained Categories:} The first issue for building a new benchmark dataset is the fine-grained categories of web images we shall collect from the Internet. In the literature, there are three famous manually labeled fine-grained datasets, \ie, FGVC-Aircraft~\cite{maji13fine-grained}, CUB200-2011~\cite{cub200-2011}, and Stanford Cars~\cite{KrauseStarkDengFei-Fei_3DRR2013}, which contain 100 types of airplanes, 200 species of birds, and 196 categories of cars, respectively. For our WebFG-496, we follow these fine-grained datasets to reuse the category labels of them as our target fine-grained categories. Furthermore, to construct a much larger and more challenging webly supervised fine-grained benchmark dataset, we explore the 5089 categories originally presented in iNat2017~\cite{van2018inaturalist} to build our WebiNat-5089. 

\textbf{Testing Images:} Since the fine-grained categories in our webly supervised datasets WebFG-496 and WebiNat-5089 come from the existing datasets as aforementioned. To save the cost of dataset construction and consider convenient comparisons with traditional fine-grained methods, we directly take the testing sets in FGVC-Aircraft, CUB200-2011, and Stanford Cars as the testing data for our WebFG-496. For WebiNat-5089, the validation set of iNat2017 is utilized as the testing data.

\textbf{Web Sources:} As pointed out by~\cite{yao2020towards}, different web sources like Google Image Search Engine (GIS), Bing Image Search Engine (BIS), Flickr, Airliners, and iNaturalist may have a significant influence on the datasets. Table~\ref{tab2} summarizes the web sources for testing images in our WebFG-496 and WebiNat-5089. To reduce the probability of overlap with the testing set along with to train webly supervised domain robust deep fine-grained models, we ultimately choose Bing Image Search Engine (BIS) as our web source of training images.

\textbf{Collecting Candidate Training Images:} Top-ranking images of image search engines tend to have relatively high accuracy and lower-ranking images usually contain more and more noises. Based on the consideration of reducing noise in the collected web images, for the 496 categories in WebFG-496, we treat each category label as a query and crawl the top 150 images from the BIS. Since some special categories are much harder to photograph than others, the natural world is extremely imbalanced~\cite{van2018inaturalist} and the realistic data obeys long-tailed distribution. For the 5089 categories in WebiNat-5089, we also treat each category label as a query but crawl as many images as possible for each query. After removing the invalid links, we obtained 71,987 images for WebFG-496 (14,817 aircraft images, 29,211 bird images, and 27,959 car images) and 1,437,483 images for WebiNat-5089, respectively. 

\textbf{Removing Broken Images:} Since our training images are directly crawled from the web, some broken images may be included. To remove these broken images, we employ the Python library \texttt{Pillow} to check each collected image and subsequently convert them to the RGB mode. Images that cannot be opened by \texttt{Pillow} or cannot be converted to RGB will be regarded as broken ones and get deleted. Table~\ref{tab2} gives a detailed number of images for WebFG-496 and WebiNat-5089 after broken images removing. 

\textbf{Removing Duplicated Images:} To remove duplicated images between training data and testing data, we take advantage of deep Convolutional Neural Network (CNN) in semantic information extraction. Our overlap removing strategy is under the assumption that images with more similar semantic information are more likely to be similar or even identical. Specifically, we first use the VGG-16~\cite{simonyan2014very} model pre-trained on ImageNet to extract the embedding feature vector for each image in both training and testing data. Here we select the feature maps of the last max-pooling layer and then perform global average pooling to cast them into a 512-d feature vector. Then, for every single test image per category, we calculate the similarity distance between this testing image and every training image. For each category, we obtain the smallest distance between training and testing data, which is denoted as $\theta$. We set an empirical threshold factor $\eta$ to scale the distance to $(1+\eta)\times \theta$ and remove the web training images which have a smaller distance than $(1+\eta)\times \theta$. In dataset constructions, the value of $\eta$ is set to be 0.01. As shown in Table~\ref{tab2}, after duplicated images removing, we ultimately obtain 53,339 training images for WebFG-496, and 1,184,520 training images for WebiNat-5089. After that, we manually check the obtained images according to the ranked distance between training and test data, and few duplicated images are left.

\textbf{Class Imbalance in WebiNat-5089:} WebiNat-5089 contains 2,101 types of plants, 1,021 kinds of insects, 289 species of reptiles, \textit{etc}. The average number of images per class for the WebiNat-5089 dataset is 232.7 while the median number is 221. The images for some categories are easy to obtain (\eg, 563 images for ``Udea Rubigalis'') while others are hard (\eg, 4 images for ``Hordnia Atropunctata''), making the extreme class imbalance a property (\ie, long-tailed) for our WebiNat-5089 dataset. Therefore, training fine-grained models from web images also needs to take the class imbalance problem into account. 

\textbf{Dataset Accuracy:} It is difficult to manually establish the accuracy of fine-grained web data, especially for WebiNat-5089 which contains over 1.1 million web training images. However, we can roughly estimate the accuracy of training data for WebFG-496 and WebiNat-5089 by random sampling. For WebFG-496, we randomly select 100 sub-categories and 50 images for each sub-category. For WebiNat-5089, we randomly select 200 sub-categories and 50 images for each sub-category. 
Finally, the roughly estimated accuracy of training data is shown in Table~\ref{tab5-2}.

\begin{table}[h]
	\small
	\centering
	\renewcommand{\arraystretch}{1.1}
	\caption{Rough accuracy of training data estimated by random sampling for WebFG-496 and WebiNat-5089.}
	\vspace{0.1cm}
	\resizebox{8.5cm}{!}{
		\begin{tabular}{c|c|c|c|c}
			\hline			
			\multirow{2}{*}{\textbf{\:\:Dataset\:\:}} & \multirow{2}{*}{\textbf{\:\:WebiNat-5089\:\:}}   & \multicolumn{3}{c}{\textbf{WebFG-496}} \\
			\cline{3-5} &                                      &  \:\:Web-aircraft\:\:   &   \:\:Web-bird\:\:   &   \:\:Web-car\:\:   \\
			\hline		
			\textbf{Accuracy}	&  36\%   &  73\%            &  65\%           & 67\%         \\
			\hline	
	\end{tabular}}
	\label{tab5-2}
	\vspace{-1em}
\end{table}

\begin{figure*}[t]
	\centering
	\includegraphics[width=0.99\textwidth]{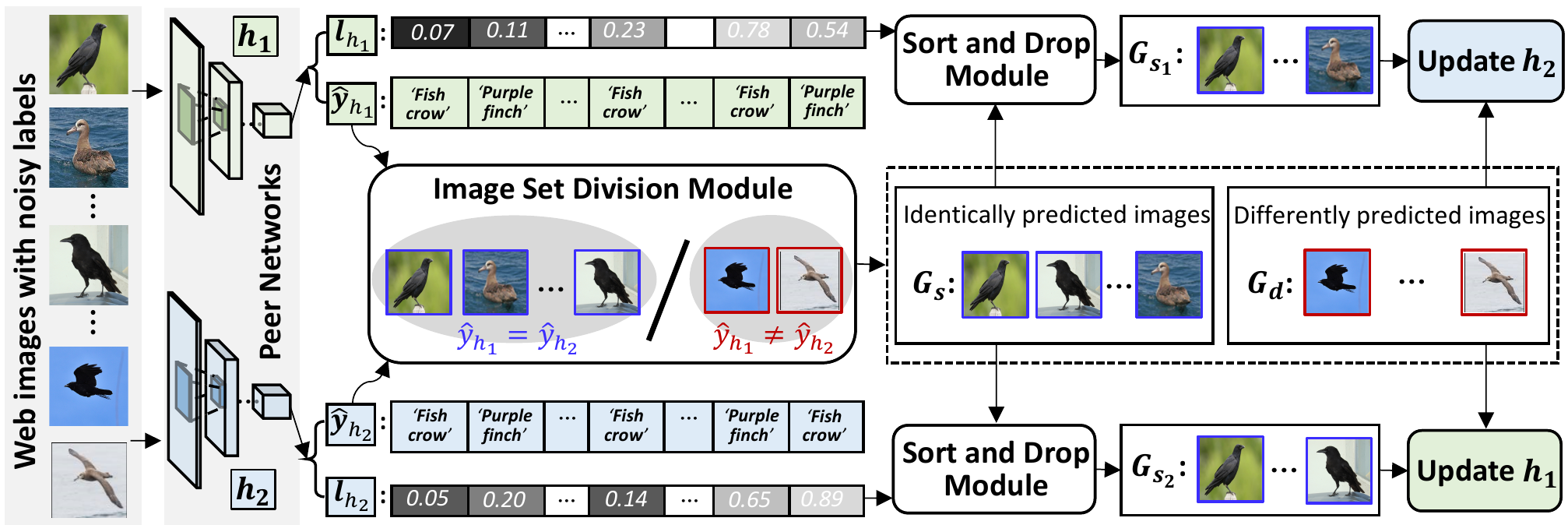}
	\caption{\small{Architecture of our Peer-learning model. The input is a mini-batch of web images. Each network in $ h_1 $ and $ h_2 $ individually feeds forward data to separately predict the labels, based on which the input data is split into two sets $ G_s $ (instances with identical predictions) and $ G_d $ (instances with different predictions). Then, $ h_1 $ and $ h_2 $ individually sort and fetch small-loss instances in $ G_s $ as the useful knowledge $ G_{s_1} $ and $ G_{s_2} $. Subsequently, $ h_1 $ updates its parameters using $ G_d $ and $ G_{s_2}, $ while $ h_2 $ updates its parameters using $ G_d $ and $ G_{s_1}$.}}
	\label{fig2}
	\vspace{-0.2cm} 
\end{figure*}

\section{Proposed Peer-learning Network}

\textbf{Training strategy} 
As shown in Fig.~\ref{fig2}, our framework consists of two networks $h_{1}$ and $h_{2}$, which mutually communicate useful information for boosting the final performance. Specifically, suppose we have a mini-batch of data $G=\{(x_i, y_i )\}$, where $y_{i}$ is the label with noise of the image $x_{i}$. $h_{1} $ and $ h_{2} $ first separately predict the labels $\{\hat{y}_{i,h_{1}}\}$ and $\{\hat{y}_{i,h_{2}}\}$ of $\{x_{i}\}$, based on which $G$ is divided into $G_s=\{(x_{k},y_{k})\in G|\hat{y}_{k,h_{1}}=\hat{y}_{k,h_{2}}\}$ (instances with identical predictions) and $G_d=\{(x_{k},y_{k})\in G|\hat{y}_{k,h_{1}}\neq \hat{y}_{k,h_{2}}\}$ (instances with different predictions). Inspired by~\cite{decoulping}, we treat $G_{d}$ as ``hard examples'', which can benefit the training of $ h_{1} $ and $ h_{2} $. To alleviate the negative effects caused by label noises in $G_{d}$, we explore ``useful knowledge'' in $G_s$ by selecting a small proportion of instances $G_{s_{1}}$ and $G_{s_{2}}$, according to the losses computed by $h_{1}$ and $h_{2}$, respectively. $G_{s_{i}}$ ($i\in\{1,2\}$) consists of instances that have the top $(1-d(T))$ smallest training losses by using $h_{i}$, and is concretely defined as:
\begin{equation}\label{eq2}\small
	\begin{matrix}
		G_{s_{i}} = \arg\min_{\tilde{G}_{s} \subset G_{s}: | \tilde{G}_{s}| \ge (1-d(T))|G_s|} \sum_{(x_{j},y_{j})\in \tilde{G}_{s}}\mathcal{L}_{h_{i}}(x_{j},y_{j}),
	\end{matrix}
\end{equation}
where $\mathcal{L}_{h_{i}}(x_{j},y_{j})$ is the training loss of instance $x_{j}$ computed by $h_{i}$ ($i\in\{1,2\}$), and $|G_{s}|$ indicates the number of elements in $G_{s}$.
Particularly,
\begin{equation}\label{eq2-2}\small
	\begin{matrix}
		d(T) = \xi \cdot \min\{\frac{T}{T_k}, 1\},
	\end{matrix}
\end{equation}
is the drop rate that dynamically controls $|G_{s_{i}}|$, where $\xi$ is the maximum drop rate, and $ T_k $ is the number of epochs after which $d(T)$ is no longer updated. The motivation lies in that we attempt to utilize more images for training at the beginning by setting $ d(T) $ to a small value. Thereafter, $d(T)$ gradually increases, so that we only select correctly labeled instances with sufficiently high confidence, and drop noisy ones before the networks memorize them.

After obtaining $G_{d}$ and $G_{s_{i}}$ ($i\in\{1,2\}$), we treat $G_{d}\cup G_{s_{1}}$ as the ``useful knowledge'' to train $h_{2}$, provided by its peer network $h_{1}$. Similarly, $G_{d}\cup G_{s_{2}}$ is adopted to train $h_{1}$. The parameters $\theta_{h_{i}}$ of the network $h_{i}$ ($i\in\{1,2\}$) are updated by using the gradient $\nabla \mathcal{L}_{h_{i}}$ with a learning rate $\lambda$:
\begin{equation}\label{eq4}
	\begin{matrix}
		\theta_{h_{1}} \leftarrow \theta_{h_{1}} - \lambda\cdot \sum_{(x_{i},y_{i})\in G_{d}\cup G_{s_{2}} } \nabla \mathcal{L}_{h_{1}}(x_{i},y_{i}),
	\end{matrix}
\end{equation}
\begin{equation}\label{eq5}
	\begin{matrix}
		\theta_{h_{2}} \leftarrow \theta_{h_{2}} - \lambda\cdot \sum_{(x_{i},y_{i})\in G_{d}\cup G_{s_{1}} } \nabla \mathcal{L}_{h_{2}}(x_{i},y_{i}).
	\end{matrix}
\end{equation}
Either network in our method learns ``useful knowledge'' from its peer network, which can correct errors caused by instances in $ G_d $ with label noises. Therefore, we denote our algorithm as ``Peer-learning''. Through mutually communicating ``useful knowledge'', both the performance of $h_{1}$ and $h_{2}$ can be improved.

\textbf{Fine-grained module}
It is worth noting that the training strategy is independent of the backbone network structure $h_{1}$ and $h_{2}$. In the Peer-learning network, we realize fine-grained recognition by using existing fine-grained modules (\eg, B-CNN \cite{bilinear}, NTS-Net \cite{yang2018}, H-B-Pooling \cite{bilinearpooling}, DFL-CNN \cite{wang2018}, OPAM \cite{OPAM}, and Fast-MPN-COV \cite{li2018}) as the backbone network structure for $h_{1}$ and $h_{2}$.

\begin{table*}[t]\small
	\centering
	\renewcommand{\arraystretch}{1.1}
	\caption{\small{The comparison of classification accuracy (\%) for benchmark methods and webly supervised baselines (Decoupling, Co-teaching, and our Peer-learning) on the WebFG-496 dataset.}}
	\vspace{0.1cm}
	\begin{tabular}{c|r|p{1.8cm}<{\centering}|p{1.8cm}<{\centering}|p{1.8cm}<{\centering}|p{1.8cm}<{\centering}|p{1.8cm}<{\centering}}
		\hline
		\multirow{2}{*}{\textbf{Type}}  &  \multirow{2}{*}{\textbf{Method\qquad}} & \multirow{2}{*}{\textbf{Backbone}}   & \multicolumn{4}{c}{\textbf{WebFG-496}} \\		
		\cline{4-7}  &                     &           &  Web-Bird   &  Web-Aircraft  &  Web-Car  &  Average   \\
		\hline
		
		\multirow{6}{*}{Benchmarks}			
		& VGG-16~\cite{simonyan2014very}    & -         & 66.34       & 68.38          &  61.62    &  65.45     \\
		& VGG-19~\cite{simonyan2014very}    & -         & 67.69       & 70.99          &  67.21    &  68.63     \\
		& ResNet-50~\cite{resnet}           & -         & 64.43       & 60.79          &  60.64    &  61.95     \\
		& ResNet-101~\cite{resnet}          & -         & 66.74       & 63.46          &  65.51    &  65.24     \\
		& GoogLeNet~\cite{googlenet}        & -         & 66.01       & 66.02          &  65.87    &  65.97     \\
		& B-CNN~\cite{bilinear}             & VGG-16    & 66.56       & 64.33          &  67.42    &  66.10     \\
		\hline  
		\multirow{3}{*}{Webly}
		& Decoupling~\cite{decoulping}	    & B-CNN     & 70.56       & \textbf{75.97} &  75.00    &  73.84     \\
		& Co-teaching~\cite{coteaching}     & B-CNN     & 73.85       & 72.76          &  73.10    &  73.24     \\
		& \textbf{Peer-learning}            & B-CNN     &\textbf{76.48}& 74.38         &  \textbf{78.52}  &  \textbf{76.46}    \\
		\hline
	\end{tabular}\\
	\label{tab3}
\end{table*}

\begin{table*}[t]
	\begin{minipage}{0.46\linewidth}
		\renewcommand{\arraystretch}{1.1}
		\centering
		\caption{The comparison of classification accuracy (\%) for Peer-learning by using different fine-grained backbone modules on the WebFG-496 dataset.}
		\vspace{0.1cm}
		\resizebox{6.2cm}{!}{
			\begin{tabular}{c|r|c}
				\hline
				\textbf{Method}    &   \textbf{Backbone\quad\:\:}               &   \textbf{WebFG-496}      \\
				\hline
				\multirow{5}{*}{\textbf{Peer-learning}}
				& NTS-Net \cite{yang2018}                    & 78.17                        \\
				& H-B-Pooling \cite{bilinearpooling}                        & 77.06                        \\
				& DFL-CNN \cite{wang2018}                           & 76.03                        \\
				& OPAM \cite{OPAM}                           & 75.32                        \\
				& Fast-MPN-COV \cite{li2018}                           & 77.10                        \\	
				
				\hline        
		\end{tabular}}
		\label{tab4}
	\end{minipage}
	\hspace{1cm}
	\begin{minipage}{0.46\linewidth} 
		\centering
		\renewcommand{\arraystretch}{1.1}
		\caption{The comparison of classification accuracy (\%) of benchmarks and our proposed webly supervised baseline Peer-learning on the WebiNat-5089 dataset.}
		\vspace{0.1cm}
		\resizebox{8cm}{!}{
			\begin{tabular}{c|r|c|c}
				\hline
				\textbf{Type}    &   \textbf{Method\quad\:\:}              &   \textbf{Backbone}  &   \textbf{WebiNat-5089}      \\
				\hline
				\multirow{3}{*}{Benchmarks}
				& VGG-16~\cite{simonyan2014very}  & -                    & 44.77                        \\
				& GoogLeNet~\cite{googlenet}      & -                    & 39.71                        \\
				& ResNet-50~\cite{resnet}         & -                    & 48.23                        \\
				\hline
				\multirow{1}{*}{Webly} 
				& \textbf{Peer-learning}          & ResNet-50            & \textbf{54.56}                             \\               
				\hline        
		\end{tabular}}
		\label{tab5}
	\end{minipage}
\end{table*}

\textbf{Why is our approach effective?} 
Since instances in $ G_d $ include biased predictions (at least one network gives a wrong prediction), updating the parameters only using instances in $ G_d $ may result in errors and these errors will be directly transferred back to itself in the next mini-batch. These errors will be increasingly accumulated and previous works, like Decoupling~\cite{decoulping} and MentorNet~\cite{mnet}, cannot handle them explicitly. In our approach, as two networks have different learning abilities, they can communicate ``useful knowledge'' (instances with a small loss in $ G_s $) with their peer network to update the parameters. Through these exchange procedures, errors induced by updating with $ G_d $ can be identified and reduced by peer networks mutually. This is the reason why our approach is more robust to previous works like Decoupling and MentorNet. On the other side, compared to methods that update the model using only ``clean'' samples (\eg, Co-teaching~\cite{coteaching}), the advantage of our approach is that we can also leverage the so-called ``noisy'' samples to promote model optimization. The motivation is that, in the group $ G_d $, there are still many correctly classified instances. Thus, we can still learn ``useful knowledge'' from these instances to improve the model performance as long as the label noise is conquered.

\textbf{Why do we need to update the neural networks with $G_d$?}
For each instance in $G_d$, at least one network in $ h_1 $ and $ h_2 $ has been given the wrong predictions. These wrong predictions give rise to an updated model that can promisingly result in better classification performance. However, updating neural networks $ h_1 $ and $ h_2 $ with $G_d$ also brings some errors. For example, the labels of instance $x$ predicted by $ h_1 $ and $ h_2 $ are $+1$ and $-1$, respectively. If the ground truth label of instance $x$ is $+1$, then $ h_2 $ treats $x$ as a wrong prediction and updating the parameters will promote the robustness of $ h_2 $. At the same time, $ h_1 $ correctly predicts the label of instance $x$, but wrongly regards it has been given an error prediction and so updates its parameters by treating $x$ as a wrong prediction will bring errors to $ h_1 $. What's worse, these errors will be transferred back and increasingly accumulated.

\textbf{Why can ``useful knowledge'' correct the errors induced by $G_d$?}
Memorization effects~\cite{motivation2017} indicate that, on noisy data sets, deep neural networks will first learn the clean and easy patterns in the initial epochs. Thus, they can filter out noisy instances through their loss values at the beginning of training. However, as the number of epochs increases, deep neural networks will eventually overfit on noisy labels. Our key idea is to drop these noisy labels before they are memorized. In our approach, we dynamically control the drop rate of noisy labels by the parameter $ d(T) $. As the number of epoch $ T $ increases, we gradually increase the drop rate $ d(T) $, so that we can keep ``clean'' instances and drop noisy labels before the neural networks memorize them. Intuitively, small-loss instances in $ G_s $ (easy examples) are more likely to be that correctly classified. On the other hand, networks $ h_1 $ and $ h_2 $ have different decision boundaries and thus have different abilities to learn. When training on noisy web images, they can have different abilities to filter out noisy labels. We cross-update neural networks using ``useful knowledge'' from each other. Through these exchange procedures of useful knowledge, different types of errors induced by $ G_d $ can be mutually identified and reduced by peer networks.    
This process is similar to ``peer-review''. When students check their own papers, it is very hard for them to find errors because they have some personal bias for the answers. Fortunately, they can ask peer classmates to review their papers. 

\begin{table*}[t]\small
	\centering
	\renewcommand{\arraystretch}{1.1}
	\caption{\small{The comparison of classification accuracy (\%) for using (or not) web images as the data augmentation. Improvement over the baseline model is reported as ($\Delta$).}}
	\vspace{0.1cm}
	\begin{tabular}{p{2.2cm}<{\centering}|p{5cm}<{\centering}|p{1.8cm}<{\centering}|p{1.8cm}<{\centering}|p{2.2cm}<{\centering}}			
		\hline
		\textbf{Testing data}   &   \textbf{Training data}   & \textbf{Backbone}   &   \textbf{ACA (\%)}  &   \textbf{Improvement}   \\
		\hline			
		\multirow{2}{*}{FGVC-Aircraft}     	 &  FGVC-Aircraft                  &    \multirow{2}{*}{VGG-16}         &  84.8  & \multirow{2}{*}{$\Delta$ \textbf{3.6}}\\ 
		&  Web-aircraft + FGVC-Aircraft   &                                    &  88.4  & \\ 
		\hline
		\multirow{2}{*}{CUB200-2011}		
		&  CUB200-2011                 &    \multirow{2}{*}{VGG-16}      &  77.7  & \multirow{2}{*}{$\Delta$ \textbf{8.0}}\\ 
		&  Web-bird + CUB200-2011      &      &  85.7  &\\
		\hline
		\multirow{2}{*}{Stanford Cars}		
		&  Stanford Cars               &    \multirow{2}{*}{VGG-16}      &  85.6  &\multirow{2}{*}{$\Delta$ \textbf{6.8}}\\ 
		&  Web-car + Stanford Cars     &      &  92.4  &\\			
		\hline
	\end{tabular}
	\label{tab6}
	\vspace{-0.5em}
\end{table*}

\begin{figure*}[t]
	\centering
	\includegraphics[width=0.99\textwidth]{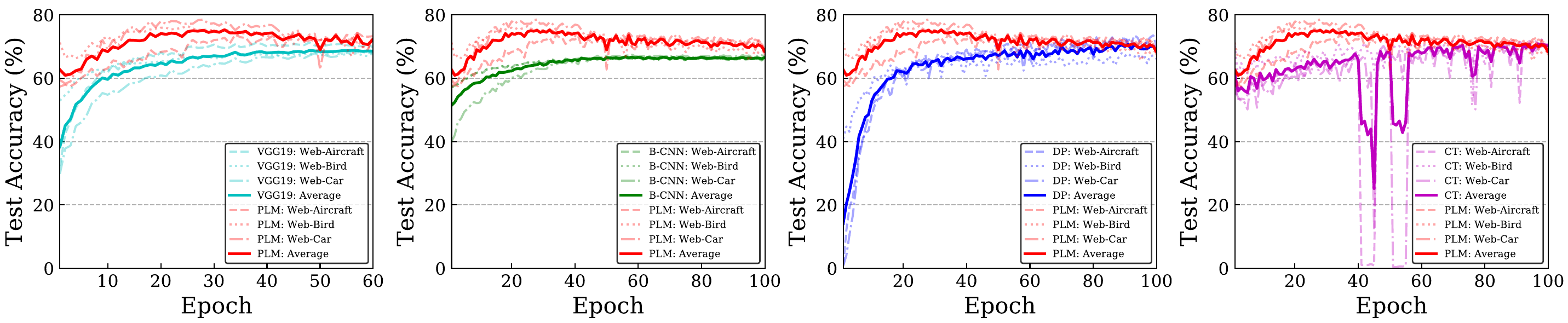}
	\caption{\small{Comparisons of classification accuracy (\%) among our Peer-learning model (PLM), VGG-19, B-CNN, Decoupling (DP), and Co-teaching (CT) on sub-datasets Web-aircraft, Web-bird, and Web-car in WebFG-496. The value on each sub-dataset is plotted in the dotted line and the average value is plotted in solid line. Note that the classification accuracy is the result of the second stage in the two-stage training strategy. Since we have trained 60 epochs in the second stage on VGG-19, we only compare the first 60 epochs in the second stage of our approach with VGG-19.}}
	\label{fig3}
	\vspace{-0.2cm}
\end{figure*} 

\section{Experiments}

\subsection{Benchmarking Two Proposed Datasets}

\textbf{Experimental Setups:} For the benchmark methods on WebFG-496, we choose Bilinear-CNN (B-CNN)~\cite{bilinear} and five deep neural networks, including VGG-16~\cite{simonyan2014very}, VGG-19~\cite{simonyan2014very}, ResNet-50~\cite{resnet}, ResNet-101~\cite{resnet}, and GoogLeNet~\cite{googlenet}. All the networks are pre-trained on ImageNet and then fine-tuned on three sub-datasets of WebFG-496. Specifically, we follow~\cite{bilinear} and adopt a two-stage training strategy. We firstly freeze the convolutional layer parameters and only optimize the last fully connected layers. Then, we optimize the parameters of all layers in the fine-tuned model. In experiments, we use SGD as optimizer. Learning rate and batch size in the first stage are set to $0.01$ and $64$, while in the second stage, they are $0.001$ and $64$. As WebiNat-5089 contains more than 1.1 million training images, it takes huge training time on such a large-scale dataset. To be efficient, for the benchmarks on WebiNat-5089, we conduct three deep networks, \ie, VGG-16~\cite{simonyan2014very}, GoogLeNet~\cite{googlenet}, and ResNet-50~\cite{resnet}. Similar to WebFG-496, we also fine-tune these deep neural networks on WebiNat-5089 but adopt a single-stage training strategy. Except for the batch size is different ($1024$ for WebiNat-5089), other parameters are the same. 

\textbf{Quantitative Results:} Experimental results for benchmarks on WebFG-496 and WebiNat-5089 are presented in Table~\ref{tab3} and Table~\ref{tab5}. As shown in Table~\ref{tab3}, we can notice that all six networks achieve reasonable fine-grained classification accuracy, which validates the reliability of WebFG-496. From Table~\ref{tab5}, we can find that the performance of three networks on WebiNat-5089 is obviously below that on WebFG-496. It indicates that WebiNat-5089 is more challenging than WebFG-496, which might be caused by the extreme class imbalance problem.

\textbf{Data Augmentation:} The widely-used FGVC benchmark datasets (\eg, CUB200-2011) all suffer from limited training data, which severely prevented the FGVC task from being sufficiently benefited from the high learning capability of deep CNNs. As a new dataset for fine-grained recognition, it is important to prove that webly supervised data can benefit fully-supervised data. To this end, we follow the semi-supervised manner and leverage web images (\ie, Web-aircraft, Web-bird, Web-car) as data augmentation w.r.t. the labeled training data for training deep FGVC models. Specifically, by pre-training VGG-16 with our webly fine-grained datasets, the improvements over baselines without leveraging webly data are shown in Table~\ref{tab6}.

\begin{figure*}[t]
	\begin{minipage}{0.48\textwidth} 
		\centering
		\includegraphics[width=0.92\textwidth]{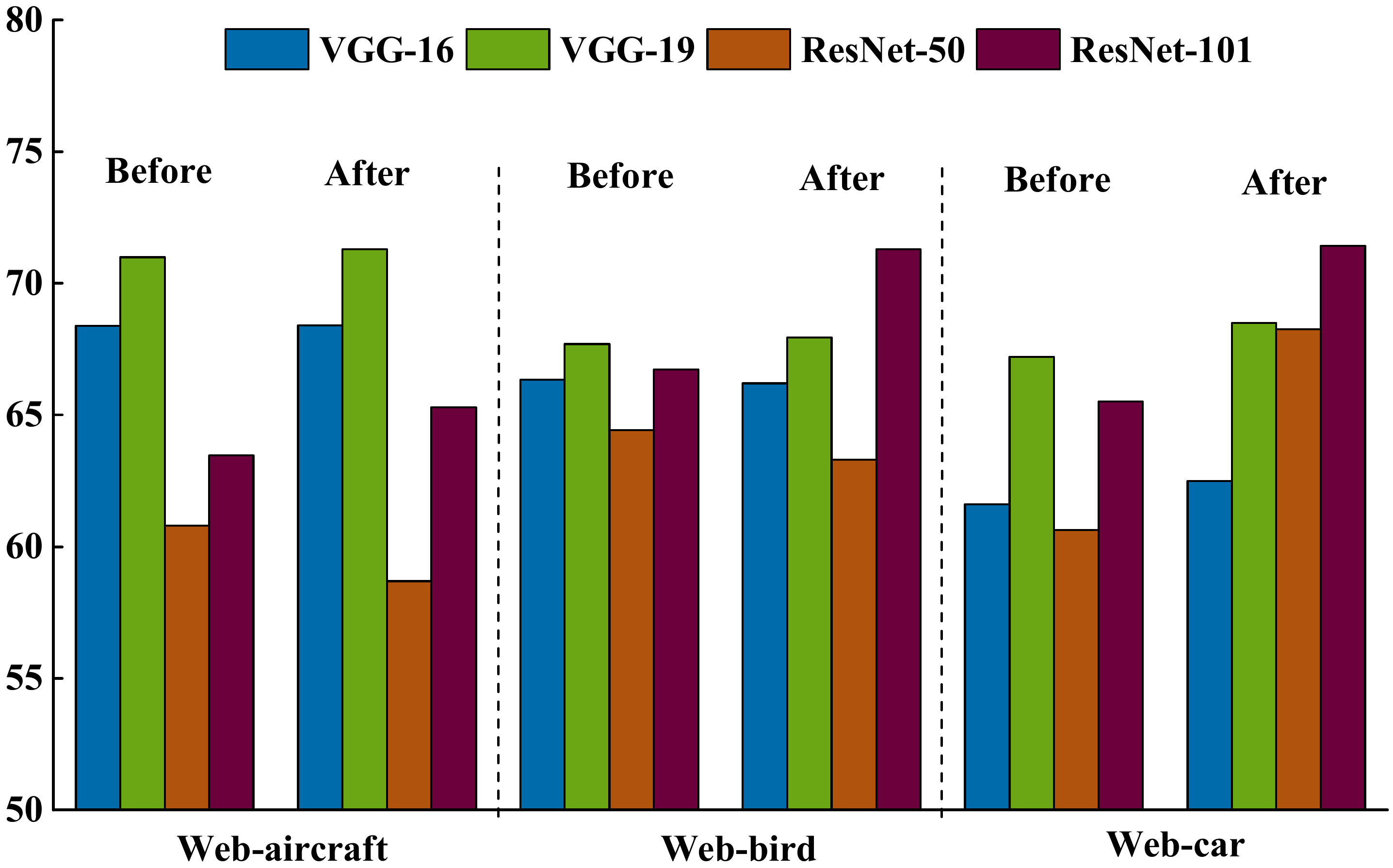}
		\caption{\small{Comparisons of classification accuracy (\%) between using (``After'') label smoothing and without using (``Before'') label smoothing on WebFG-496.}}
		\label{fig4}
		\vspace{-0.3cm}
	\end{minipage}
	\hspace{0.2cm}
	\begin{minipage}{0.49\textwidth} 
		\centering
		\includegraphics[width=0.92\textwidth]{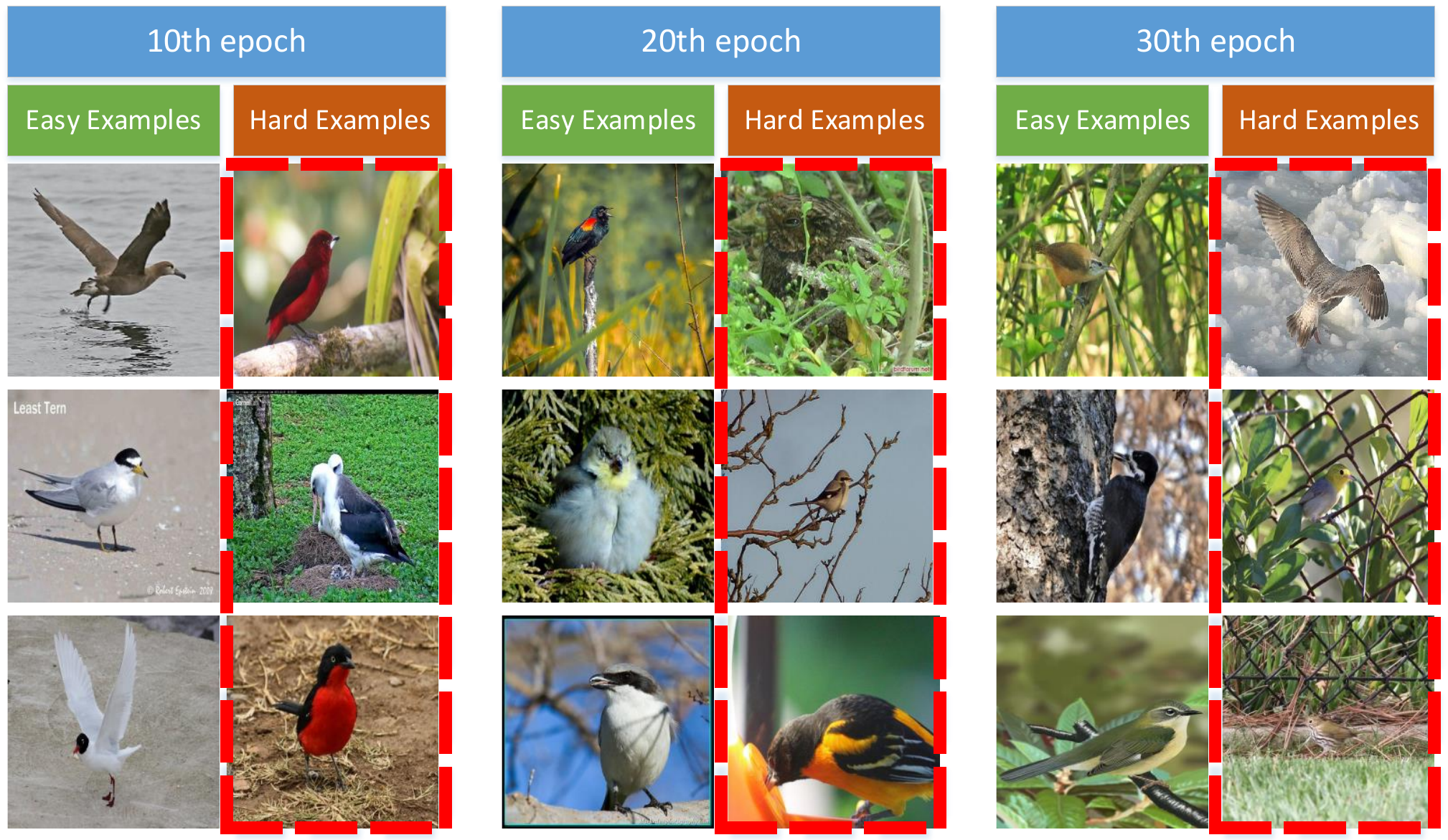}
		\caption{\small{Visualization of ``easy'' and ``hard'' examples explored by our proposed approach as optimization continues in different training epochs.}}
		\label{fig5}
		\vspace{-0.3cm}
	\end{minipage}
\end{figure*}

\textbf{Label Smoothing:} To investigate the impact of using label smoothing on webly supervised benchmark datasets, we conduct experiments on VGG-16, VGG-19, RestNet-50 and ResNet-101 by setting the smoothing parameter to be 0.1. The experimental results on WebFG-496 are given in Fig.~\ref{fig4}. From that figure, we can observe that: 1) The performances of VGG-16 and VGG-19 do not change too much after using label smoothing on three sub-datasets. 2) When using label smoothing, the performance for ResNet-50 decreases on Web-aircraft and Web-bird, but increases obviously on Web-car. 3) The performance of ResNet-101 grows incontestably on all sub-datasets after using label smoothing. 

\subsection{Peer-learning Performance}

\textbf{Implementation Details:} For WebFG-496, we use a pre-trained VGG-16~\cite{simonyan2014very} to initialize all convolutional layers of both $h_1$ and $h_2$. Network architecture of $h_1$ and $h_2$ is B-CNN~\cite{bilinear}. To avoid learning two identical networks, we perform ``Kaiming Normal Initialization'' over the fully connected layers to ensure that $ h_1 $ and $ h_2 $ have different starting points. In experiments, we set $ \xi = 0.35 $ and $ T_k = 10 $ as the default values on WebFG-496. For model training, we optimize by Adam and also adopt a two-stage training strategy. Learning rate and batch size in the first stage are set to $10^{-3}$ and $64$, while in the second stage, they are $10^{-4}$ and $32$. For WebiNat-5089, we take pre-trained ResNet-50~\cite{resnet} to initialize $ h_1 $ and $ h_2 $. Network architecture of $ h_1 $ and $ h_2 $ is also ResNet-50. We set $ \xi = 0.3 $ and $ T_k = 10 $ on WebiNat-5089. For model training, a single-stage training strategy is performed to save training time. Learning rate, batch size, and training epoch are set to $10^{-3}$, $160$, and $60$, respectively.

\textbf{Baselines:} On WebFG-496, we compare our Peer-learning with two webly supervised state-of-the-art baseline approaches: Decoupling~\cite{decoulping} and Co-teaching~\cite{coteaching}. To be fair, we replace the basic network in these two methods and use the same backbone network B-CNN~\cite{bilinear} as ours. For all other parameters like batch sizes, learning rates, weight decay, training epochs, and drop rates, we all set the same as our Peer-learning. In addition, we replace the backbone network of B-CNN \cite{bilinear} in Peer-learning and use SOTA fine-grained modules NTS-Net \cite{yang2018}, H-B-Pooling \cite{bilinearpooling}, DFL-CNN \cite{wang2018}, OPAM \cite{OPAM}, and Fast-MPN-COV \cite{li2018} for comparison.
On WebiNat-5089, we directly compare Peer-learning with benchmarks. The reason is that training a deep model on such a large-scale dataset needs more than 240 hours under the computing configuration: 4 V100 GPU (32G) cards with a batch size 160.

\textbf{Quantitative Results:} Experimental results are shown in Table~\ref{tab3}, Table~\ref{tab4}, Table~\ref{tab5}, and Fig.~\ref{fig3}. As shown in Table~\ref{tab3}, we can notice that our Peer-learning greatly improves the performances of webly supervised methods. Compared to Co-teaching, our Peer-learning can not only leverage the useful knowledge in $G_{s1}$ and $G_{s2}$ to update the network parameters, but can also use the noisy data in $G_d$ for promoting the parameter optimization. Compared to Decoupling, each network in our approach can learn useful knowledge from its peer network to correct the errors when updating. Thus, the errors increasingly accumulated in Decoupling can be greatly reduced in our approach. From Table~\ref{tab3} and Table~\ref{tab5}, we can observe that our approach performs much better than baseline networks. Compared to~\cite{simonyan2014very,resnet,googlenet}, our approach maintains two networks that can learn useful knowledge from each other. 
From Table~\ref{tab4}, we can notice that our Peer-learning can combine with different SOTA fine-grained modules and achieve leading results.
By observing Fig.~\ref{fig3}, we can find that our approach is not only better than other methods, but also much faster to achieve model optimization. The explanation is that our approach can gradually drop the noisy images and select useful samples for training models.

\textbf{Visualization of Different Epochs:} Fig.~\ref{fig5} visualizes ``easy'' and ``hard'' examples explored by our method in the 10-\textit{th}, 20-\textit{th} and 30-\textit{th} epoch, respectively. From that figure, we can notice that with the grows of training epochs, the recognition ability of our model is increasing. For example, compared to the 10-\textit{th} epoch, our model in the 30-\textit{th} epoch can recognize more complex examples. 

\section{Conclusion}

In this work, we studied the problem of fine-grained recognition through noisy web images. To be specific, we first constructed two benchmark webly supervised fine-grained datasets WebFG-496 and WebiNat-5089. Then we proposed a novel method, termed Peer-learning, by simultaneously training two deep neural networks, both of which exploit and mutually communicate useful knowledge from noisy web images. Extensive experiments showed that our approach achieved state-of-the-art performance, compared to existing works. In the future, we plan to investigate the impact of class imbalance and design a more robust fine-grained model that takes class imbalance into account.

\section*{Acknowledgments}

This work was supported by the National Natural Science Foundation of China (No. 61976116, 61772256, and 61772256), the Fundamental Research Funds for the Central Universities (No. 30920021135 and 30920041111), and the CAAI-Huawei MindSpore Open Fund (CAAIXSJLJJ-2020-022A).

\end{document}